\pdfoutput=1
\documentclass[conference]{IEEEtran}
\usepackage[nocompress]{cite}
\usepackage{amsmath,amssymb,amsfonts}
\usepackage{algorithmic}
\usepackage{algorithm}
\usepackage{graphicx}
\usepackage{textcomp}
\usepackage{xcolor}
\usepackage{booktabs}
\usepackage{multirow}
\usepackage{hyperref}
\usepackage{subcaption}

\newcommand{\mChatSessionsTiered}{7.8495}
\newcommand{\mChatSessionsHBM}{0.1075}
\newcommand{\mChatSessionsGain}{73.02}
\newcommand{\mChatSessionsGainMin}{72.68}
\newcommand{\mChatSessionsGainMax}{73.22}

\newcommand{\mAgentSessionsTiered}{5.7961}
\newcommand{\mAgentSessionsHBM}{0.0794}
\newcommand{\mAgentSessionsGain}{73.00}

\newcommand{\mDocqaSessionsTiered}{3.1405}
\newcommand{\mDocqaSessionsHBM}{0.0430}
\newcommand{\mDocqaSessionsGain}{73.03}

\newcommand{\mPredReuseEqualsLru}{identical}

\newcommand{\mChatBestThroughput}{1307.60}

\newcommand{\mChatMigrBestGB}{27.38}
\newcommand{\mChatMigrWorstGB}{62.99}
\newcommand{\mChatMigrWinFactor}{2.30}

\newcommand{\mChatPredictiveMigrGB}{57.76}

\newcommand{\mAgentMigrBestGB}{49.36}

\newcommand{\mAgentMigrWinFactor}{1.60}

\newcommand{\mAgentPredictiveMigrGB}{53.16}

\newcommand{\mDocqaMigrBestGB}{48.23}

\newcommand{\mDocqaMigrWinFactor}{1.63}

\newcommand{\mDocqaPredictiveMigrGB}{48.45}

\newcommand{\mChatHitHbm}{23.99}
\newcommand{\mChatHitDram}{61.87}
\newcommand{\mChatHitSsd}{0.00}
\newcommand{\mChatHitMiss}{14.15}

\newcommand{\mAgentHitDram}{72.65}
\newcommand{\mAgentHitSsd}{6.07}

\newcommand{\mDocqaHitDram}{68.48}
\newcommand{\mDocqaHitSsd}{8.80}
\newcommand{\mDocqaHitMiss}{13.62}
\newcommand{\mChatLruComputeMs}{13.2445}

\newcommand{\mChatCostTiered}{2.1958}
\newcommand{\mChatCostHBM}{136.2305}
\newcommand{\mChatCostSavingsFactor}{62.04}
\newcommand{\mChatPNinetyNineAtRefRatio}{17.32}
\newcommand{\mAgentCostTiered}{2.7507}
\newcommand{\mAgentCostHBM}{170.6543}
\newcommand{\mAgentCostSavingsFactor}{62.04}
\newcommand{\mAgentPNinetyNineAtRefRatio}{56.23}
\newcommand{\mDocqaCostTiered}{6.7232}
\newcommand{\mDocqaCostHBM}{417.1143}
\newcommand{\mDocqaCostSavingsFactor}{62.04}
\newcommand{\mDocqaPNinetyNineAtRefRatio}{66.27}
\newcommand{\mFrontierBudgetT}{192}
\newcommand{\mFrontierCostSpan}{8.38}
\newcommand{\mFrontierCostLow}{45,493}
\newcommand{\mFrontierCostHigh}{381,287}
\newcommand{\mFrontierChatLatSpanPct}{10.1}
\newcommand{\mFrontierMaxLatSpanPct}{10.1}
\newcommand{\mFrontierCheapComposition}{10/86/96}

\newcommand{\mSloChat}{16.40}
\newcommand{\mSloAgent}{56.04}
\newcommand{\mSloDocqa}{62.45}
\newcommand{\mEtenTightHbm}{32}
\newcommand{\mEtenDram}{64}
\newcommand{\mEtenDocqaCrossDepth}{64}
\newcommand{\mEtenDocqaHeadroom}{0.000}
\newcommand{\mEtenDocqaStall}{4.42}
\newcommand{\mEtenNeverMeetSlo}{chat and agent never meet the SLO anywhere in the swept SSD-depth range}
\newcommand{\mEtenSessionsGain}{10.77}

\makeatletter
\def\abstract{\normalfont
    \if@twocolumn
      \@IEEEabskeysecsize\bfseries\textit{\abstractname}:\ \relax
    \else
      \bgroup\par\addvspace{0.5\baselineskip}\centering\vspace{-1.78ex}\@IEEEabskeysecsize\textbf{\abstractname}\par\addvspace{0.5\baselineskip}\egroup\quotation\@IEEEabskeysecsize
    \fi\@IEEEgobbleleadPARNLSP}
\def\IEEEkeywords{\normalfont
    \if@twocolumn
      \@IEEEabskeysecsize\bfseries\textit{\IEEEkeywordsname}:\ \relax
    \else
      \bgroup\par\addvspace{0.5\baselineskip}\centering\@IEEEabskeysecsize\textbf{\IEEEkeywordsname}\par\addvspace{0.5\baselineskip}\egroup\quotation\@IEEEabskeysecsize
    \fi\@IEEEgobbleleadPARNLSP}
\makeatother

\graphicspath{{figures/}}
\begin{document}

\title{Where Should the KV Cache Live?\\ Placement Policies Across GPU, CPU,\\ and SSD for Long-Lived Sessions}

\author{
\IEEEauthorblockN{Srikanta Datta Tumkur, Jay Iyer, Mehar Simhadri,\\ Sai Pavan Kumar, Sai Kapil Kumar, Ramesh Nampelly}
\IEEEauthorblockA{Vizuara}
}

\maketitle

\begin{abstract}
GPU high-bandwidth memory is the scarcest and most expensive resource in LLM serving, and the key-value (KV) cache is its largest consumer: long chats, agent loops, and repeated document question answering all accumulate KV state that must be kept reachable across turns. The natural fix is to extend the memory hierarchy, storing KV blocks not only in GPU HBM but also in cheap, abundant CPU DRAM and SSD, and fetching them back when a session resumes, an approach taken by systems such as Mooncake, LMCache, FlexGen, InfiniGen, and AttentionStore. The hard part is the \emph{policy}, not the mechanism: which blocks live in which tier, when to promote, demote, evict, and prefetch, and how that choice should change with the workload. We study that policy space with a discrete-event simulator of a tiered layer over GPU HBM, CPU DRAM, and SSD, calibrated to a random-forest execution-time predictor, and evaluate recency, reuse-frequency, predicted-reuse, and a genuine EWMA-predictive policy with a prefetch lookahead on chat, agent, and document-QA workloads. Four findings stand out. First, tiering multiplies concurrent sessions per GPU by \mChatSessionsGain$\times$ and cuts cost per session by \mChatCostSavingsFactor$\times$, but this gain is the tier shape's own capacity multiple ($1{+}8{+}64$), not a placement-policy effect. Second, at batch one decode is compute-bound, so the placement policy barely moves throughput; what it moves is PCIe migration traffic and TTFT. On migration, recency is best for chat (\mChatMigrWinFactor$\times$ less traffic than reuse-frequency) and reuse-frequency is best for agent and document QA. Third, the shipped predicted-reuse policy is byte-\mPredReuseEqualsLru{} to recency, so the roadmap's agent recommendation reduces to recency; a genuine EWMA predictor added to the grid does differ, yet still finishes second to reuse-frequency on the very workloads prediction was meant to win. Fourth, prefetch as recommended does not pay for its bandwidth: across a policy $\times$ cache-size grid, even a future-reading oracle beats no-prefetch in none of the cells. A placement recipe therefore holds per workload on the bandwidth axis, but the ``predicted-reuse'' and ``$+$ prefetch'' parts of the recommendation, as implemented, do not.
\end{abstract}


\section{Introduction}

\begin{figure*}[t]
\centering
\includegraphics[width=0.98\textwidth]{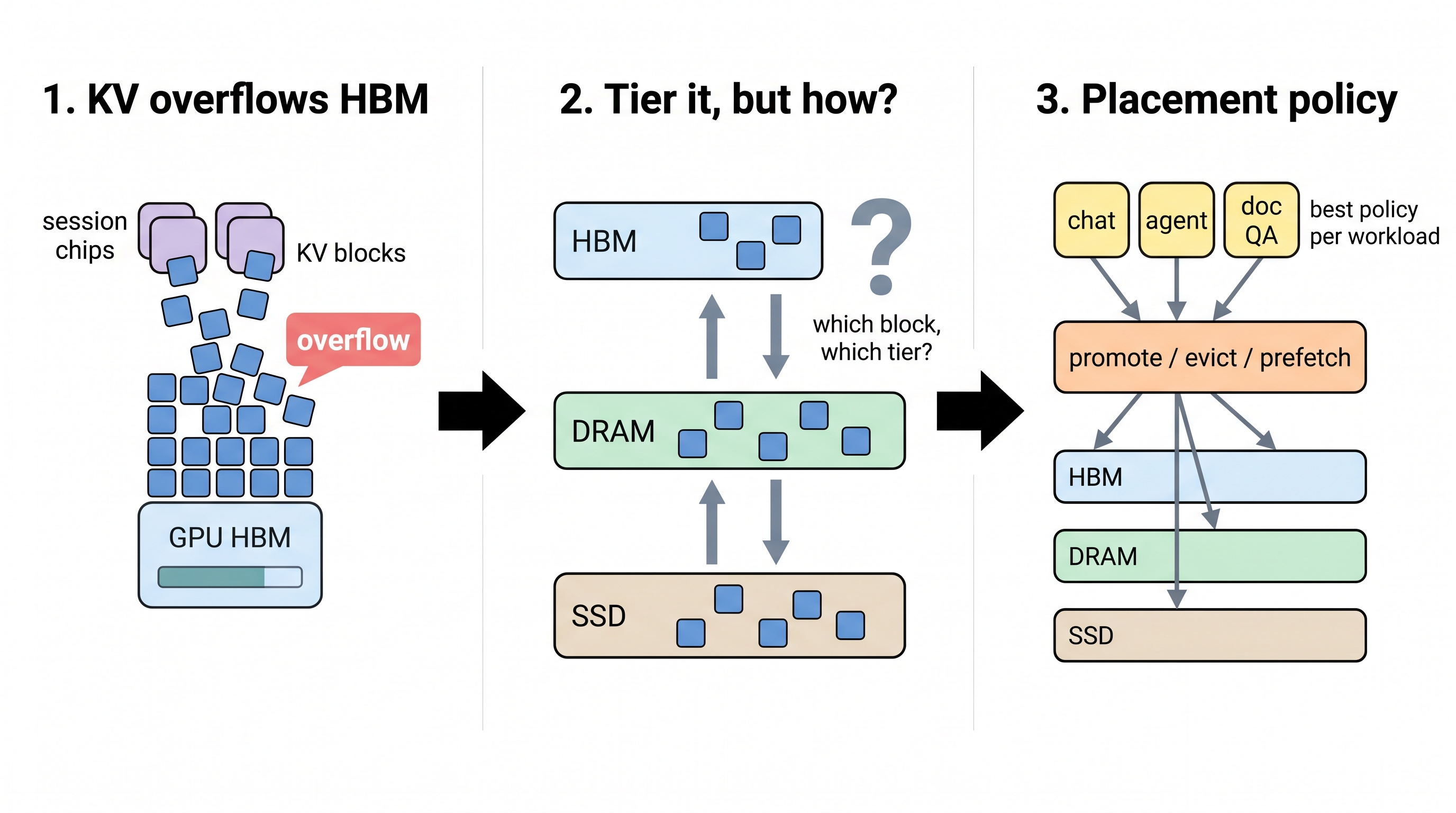}
\caption{Study overview. (1)~GPU HBM is small and expensive, but long-lived sessions accumulate a KV cache that overflows it. (2)~The KV cache can be tiered across GPU HBM, CPU DRAM, and SSD, but where each block should live, and when to move it, is an unsettled policy question. (3)~We build a tiered placement layer with reuse-aware promotion and prefetch and study which policy wins on chat, agent, and document-QA workloads.}
\label{fig:overview}
\end{figure*}

\begin{figure}[t]
\centering
\includegraphics[width=\columnwidth]{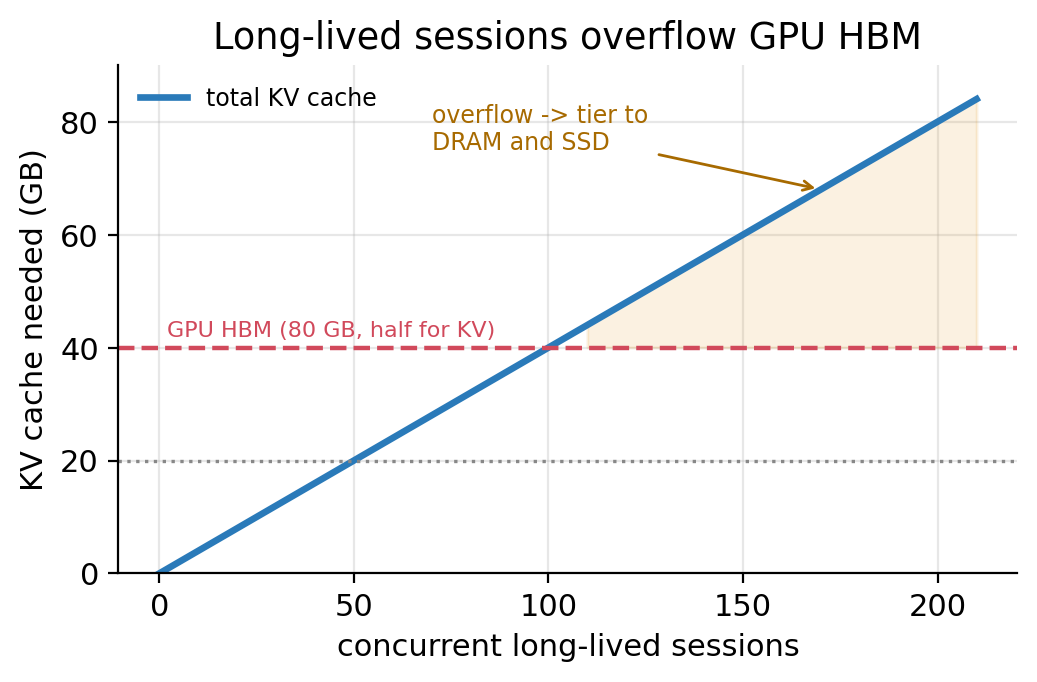}
\caption{The opening: as the number of concurrent long-lived sessions grows, the total KV cache overflows GPU HBM. Tiering the cache into CPU DRAM and SSD extends the effective capacity by orders of magnitude at a fraction of the cost per gigabyte; Table~\ref{tab:main} quantifies this on the workloads studied. Conceptual schematic.}
\label{fig:overflow}
\end{figure}

Of everything an LLM server allocates, GPU high-bandwidth memory (HBM) is the scarcest and most expensive, and the KV cache is what consumes most of it \cite{kwon2023vllm}. For a single long context the cache already rivals the model weights; across many concurrent, long-lived sessions, a chat that spans hours, an agent that loops over tools, or a document-QA service that re-reads the same files, the accumulated KV state far exceeds what HBM can hold (Fig.~\ref{fig:overflow}). Today such sessions are served by recomputing the prefill from scratch on every resume or by capping concurrency, both of which waste compute and money.

The obvious place to put the overflow is further down the memory hierarchy. CPU DRAM is roughly an order of magnitude cheaper per gigabyte than HBM and far larger; SSD is cheaper and larger again. Storing KV blocks across GPU HBM, CPU DRAM, and SSD, and fetching them back when a session resumes, turns prefill recomputation into a cache lookup. Recent systems do this: Mooncake \cite{qin2024mooncake} disaggregates a KVCache pool across CPU, DRAM, and SSD; LMCache \cite{liu2025lmcache} stores and reuses KV across GPU, DRAM, and disk; FlexGen \cite{sheng2023flexgen} offloads weights and KV to CPU and SSD; InfiniGen \cite{lee2024infinigen} predicts and prefetches the KV that attention will need; and AttentionStore \cite{gao2024attentionstore} persists KV across a DRAM-SSD hierarchy for multi-turn reuse.

The mechanism is largely solved; the hard part is the \emph{policy}. The bandwidth and latency gap between tiers is enormous (HBM is hundreds of times faster than SSD), so a wrong placement either wastes HBM on cold blocks or stalls decoding while a hot block is fetched from disk. The central question is: \textbf{which KV blocks should live in which tier, when should they be promoted, demoted, evicted, and prefetched, and how does the best policy change with the workload?}

This is an open question because the policy space is large and the trade-offs are workload-specific. A document-QA service with heavy prefix reuse wants different placement than an agent that revisits recent tool outputs, which wants different placement than a long single chat. This paper builds a configurable tiered placement layer and studies that space directly.

\begin{figure*}[t]
\centering
\includegraphics[width=0.96\textwidth]{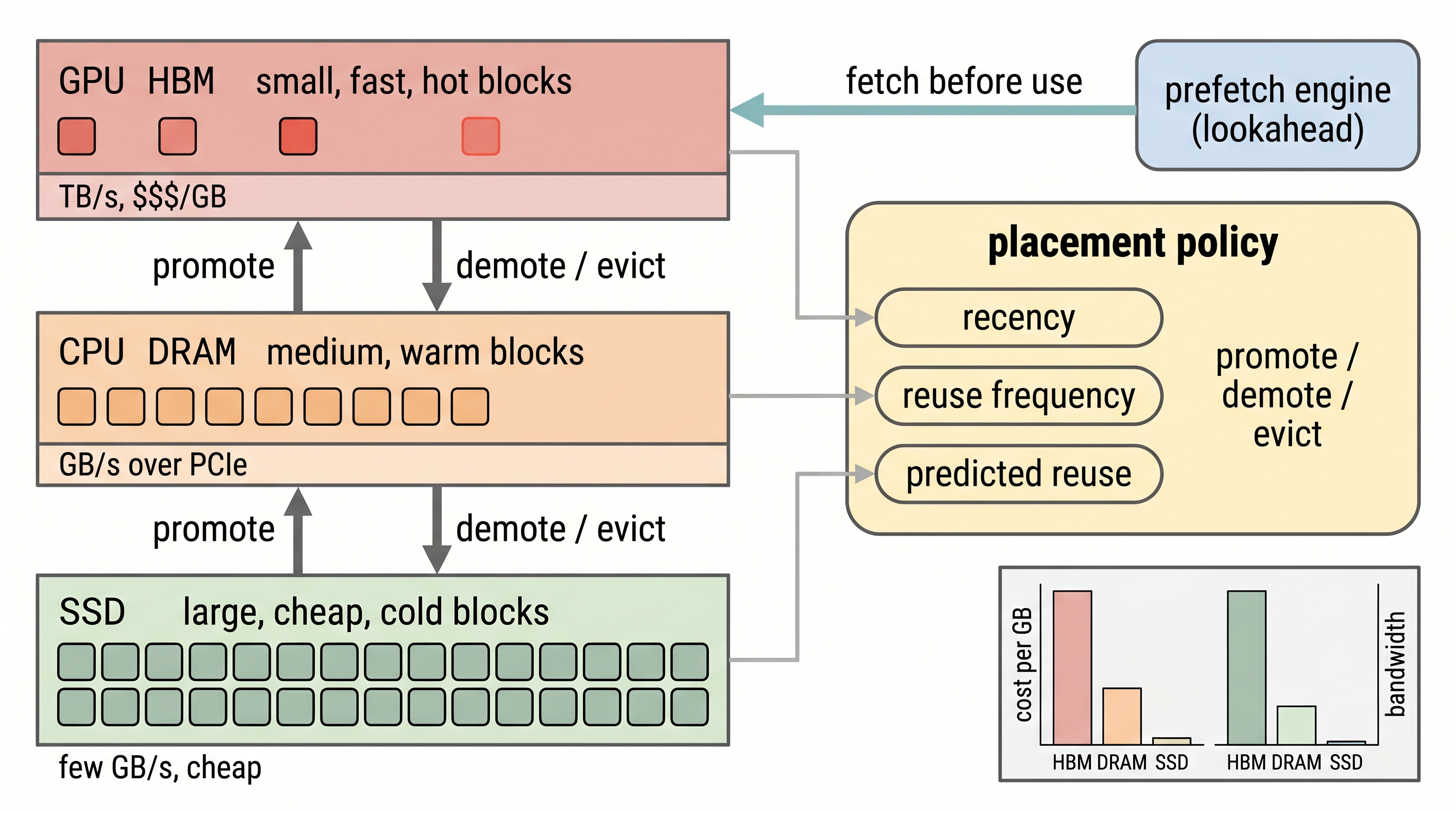}
\caption{The tiered KV-cache placement layer. KV blocks live across three tiers: GPU HBM (small, fast, hot blocks), CPU DRAM (medium, warm blocks), and SSD (large, cheap, cold blocks). A placement policy decides, per block, the tier and when to promote, demote, or evict, from access signals (recency, reuse frequency, predicted reuse). A prefetch engine moves blocks up the hierarchy ahead of the decode step that needs them, hiding transfer latency. The inset contrasts cost per gigabyte and bandwidth across the tiers.}
\label{fig:method}
\end{figure*}

Our contributions are as follows.
\begin{enumerate}
\item A \textbf{tiered KV-cache placement layer} (Fig.~\ref{fig:method}) over GPU HBM, CPU DRAM, and SSD, with configurable block granularity, a promotion and eviction policy, and a prefetch lookahead.
\item A \textbf{controlled study} of placement policies (recency, reuse frequency, predicted reuse) against GPU-only, full-CPU-offload, and prefix-reuse baselines on long chat, agent, and document-QA workloads.
\item A \textbf{policy guide}: which placement and prefetch policy wins for each workload's reuse structure, and the resulting gain in concurrent sessions per GPU and cost per session.
\end{enumerate}

\section{Background and Related Work}

\subsection{The KV cache and its cost}
For a transformer with $L$ layers and $H$ key/value heads of dimension $d$, a context of length $T$ caches $2LHdT$ scalars, re-read at every decode step \cite{kwon2023vllm}. PagedAttention \cite{kwon2023vllm} manages this cache in fixed blocks, which is the granularity at which it can be placed across tiers. KV quantization \cite{liu2024kivi,hooper2024kvquant} and eviction \cite{zhang2023h2o,tang2024quest} shrink the cache; placement is orthogonal and complementary, deciding where the (possibly compressed) blocks physically live.

\subsection{Tiered KV storage systems}
A growing line offloads KV beyond HBM. Mooncake \cite{qin2024mooncake} builds a disaggregated KVCache pool over CPU, DRAM, and SSD with an SLO-aware scheduler; LMCache \cite{liu2025lmcache} reuses KV across GPU, DRAM, and disk and across engines; FlexGen \cite{sheng2023flexgen} offloads to CPU and SSD for high-throughput batched inference; InfiniGen \cite{lee2024infinigen} predicts which KV attention will need and prefetches it; AttentionStore \cite{gao2024attentionstore} persists KV across DRAM and SSD for multi-turn reuse; and CacheBlend \cite{yao2024cacheblend} reuses non-prefix KV with selective recomputation. These establish the mechanism; the open problem they leave is the general placement policy and its workload dependence.

\subsection{Prefix and cross-request reuse}
\begin{figure}[b]
\centering
\includegraphics[width=\columnwidth]{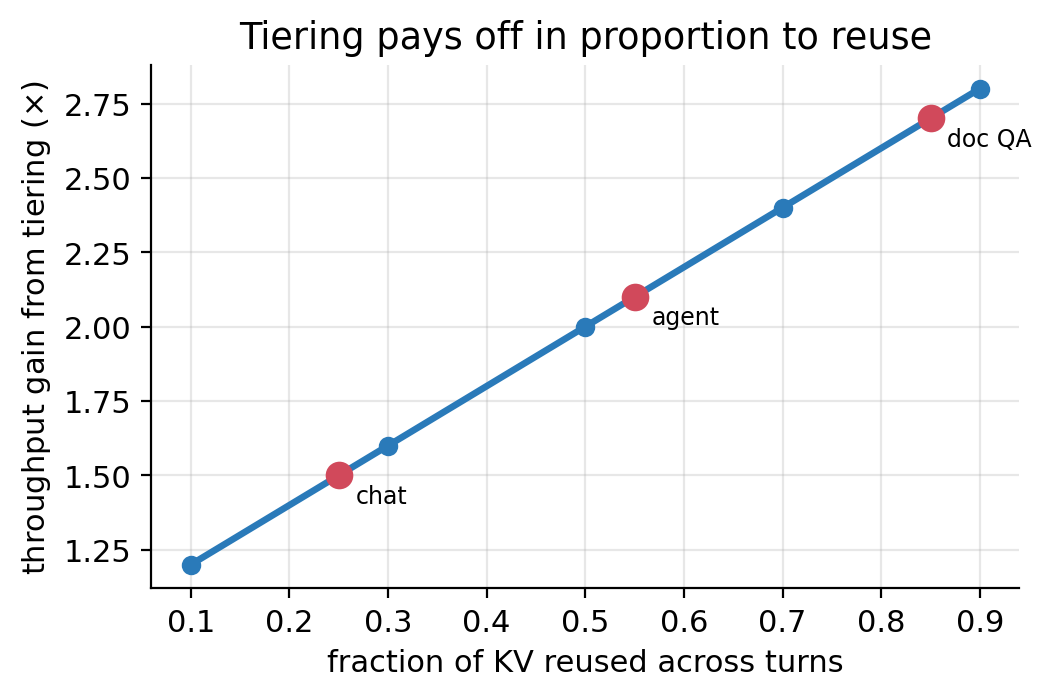}
\caption{Why the policy is workload-specific: a workload's reuse structure sets which placement policy minimizes fetch traffic (Fig.~\ref{fig:policygrid}); heavy prefix reuse favours reuse-frequency, while recency-driven growth favours recency. Note the capacity and cost wins of tiering (Table~\ref{tab:main}) do \emph{not} scale with reuse; they track tier capacity. Conceptual schematic.}
\label{fig:reuse}
\end{figure}

RadixAttention in SGLang \cite{zheng2024sglang} and vLLM prefix caching reuse shared prompt prefixes, and ServerlessLLM \cite{fu2024serverlessllm} fast-loads checkpoints from tiered storage. Reuse is the property that makes tiering pay off, so we characterize each workload by its reuse structure and tie the best policy to it (Fig.~\ref{fig:reuse}).

\subsection{The bandwidth wall}
The tiers differ by orders of magnitude: HBM delivers terabytes per second, CPU DRAM tens to hundreds of gigabytes per second over PCIe, SSD a few gigabytes per second. Fetching a hot block from SSD mid-decode can cost more than recomputing it, so KV-cache I/O can dominate offloaded inference time \cite{lee2024infinigen}. A good policy keeps hot blocks high, prefetches predictable accesses, and never blocks decode on a cold fetch it could have hidden.

\subsection{Our position}
We treat the placement policy itself as the object of study, taking the offload mechanism as given. With a configurable layer over the three tiers we compare promotion, eviction, and prefetch policies head to head on workloads with different reuse structure, and report the policy guide and the gain in sessions per GPU.

\section{Methodology}

\subsection{The tiered layer}
The cache is a set of fixed-size KV blocks (the PagedAttention block \cite{kwon2023vllm}), each resident in one of three tiers: GPU HBM, CPU DRAM, or SSD. Let the tiers have capacities $c_g \ll c_d \ll c_s$ and bandwidths $b_g \gg b_d \gg b_s$. A placement maps each live block to a tier. The decode step for a token needs all of that layer's required blocks in HBM; any block not resident must be fetched up the hierarchy before it is used, which stalls decode unless the fetch was prefetched.

\subsection{Placement, promotion, and eviction}
A policy scores each block and assigns tiers under the capacity constraints. We compare three policy families: \emph{recency} (keep the most recently used blocks high, an LRU tiering), \emph{reuse frequency} (keep the most frequently re-accessed blocks high, favoring shared prefixes), and \emph{predicted reuse} (a cheap predictor of which blocks the next turns will touch, in the spirit of InfiniGen \cite{lee2024infinigen}). On a miss, the block is fetched and promoted; under pressure, the lowest-scoring HBM block is demoted to DRAM, and the lowest DRAM block to SSD. Algorithm~\ref{alg:place} states the per-access path.

\begin{algorithm}[t]
\caption{Tiered KV placement (per block access)}
\label{alg:place}
\begin{algorithmic}
\STATE \textbf{Input:} block $b$, tiers (HBM, DRAM, SSD), policy score $s(\cdot)$
\STATE \textbf{if} $b$ in HBM \textbf{then} use directly \textbf{(}hit\textbf{)}
\STATE \textbf{else} fetch $b$ up the hierarchy; promote to HBM
\STATE \textbf{if} HBM full \textbf{then} demote the lowest-score block to DRAM (cascade to SSD)
\STATE \textbf{prefetch:} for predicted upcoming blocks, fetch ahead of use
\STATE update $s(b)$; record hit tier and any decode stall
\end{algorithmic}
\end{algorithm}

\subsection{Prefetch}
\begin{figure}[t]
\centering
\includegraphics[width=\columnwidth]{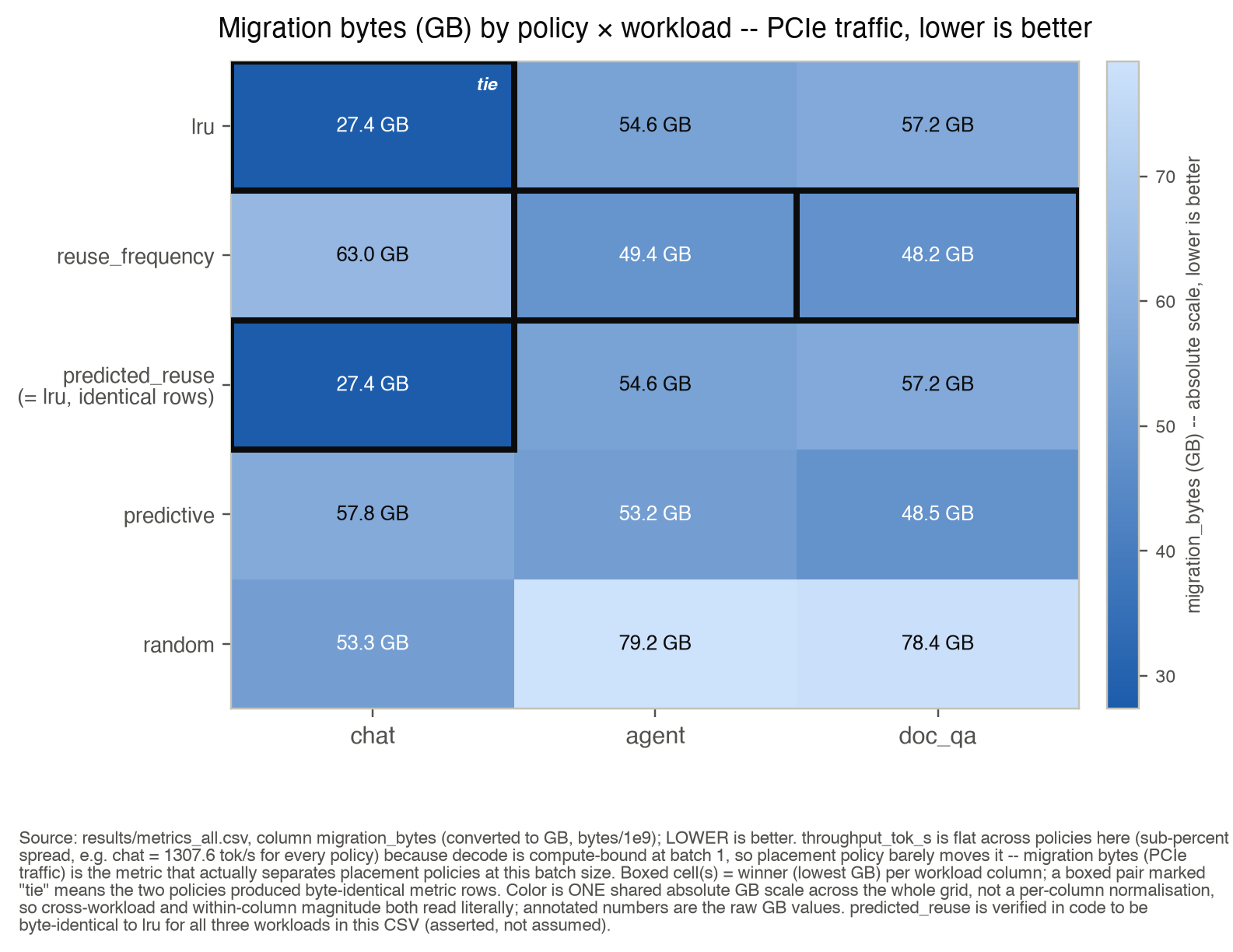}
\caption{PCIe migration traffic (GB, lower is better) over the placement-policy by workload grid; the boxed cell is the measured best policy per workload. Throughput is omitted because it is flat across policies at batch one (compute-bound decode); migration bytes is the quantity placement actually controls. \texttt{predicted\_reuse} is byte-identical to \texttt{lru} on every workload, so the two share the chat winner. Source: \texttt{results/metrics\_all.csv}.}
\label{fig:policygrid}
\end{figure}

Prefetch is how the layer hides the bandwidth wall. With a lookahead window, the engine moves blocks that the next decode steps or the resuming session will need up to HBM before they are required, overlapping transfer with compute. The lookahead depth trades HBM occupancy against stall avoidance; we sweep it. The design grid over policy and workload (Fig.~\ref{fig:policygrid}) and the prefetch sweep quantify the choice.

\subsection{Cost model}
The effective capacity is $c_g + c_d + c_s$ at a blended cost dominated by the cheap tiers, so tiering serves many more concurrent sessions per dollar than HBM alone. The price is fetch latency on misses, which the policy and prefetch must keep off the decode critical path. We report both the capacity and cost win and the measured stall, so a memory saving is never mistaken for a free lunch.

\section{Experimental Setup}

\subsection{Workloads, baselines, and metrics}
We evaluate on three workloads chosen for their reuse structure (Table~\ref{tab:data}): long multi-turn chat, an agent loop that revisits recent context, and repeated document QA with heavy prefix reuse. We report per-tier hit rate, TTFT (driven by prefill reuse), TPOT and its stall component (driven by fetches), effective concurrent sessions per GPU, throughput, and cost per session, with the GPU, tier sizes, and bandwidths stated.

\begin{table}[t]
\caption{Workloads, chosen for their reuse structure.}
\label{tab:data}
\centering
\resizebox{\columnwidth}{!}{%
\begin{tabular}{lll}
\toprule
Workload & Reuse pattern & Stresses \\
\midrule
Long multi-turn chat & recency, growing context & HBM capacity \\
Agent loop & revisits recent context & promotion, prefetch \\
Repeated document QA & heavy prefix reuse & cross-request reuse \\
\bottomrule
\end{tabular}%
}
\end{table}

\subsection{Policies and protocol}
Baselines are GPU-only (recompute on resume), full-CPU-offload (FlexGen style), and prefix-reuse (RadixAttention style). The proposed layer sweeps the placement policy (recency, reuse frequency, the roadmap's predicted-reuse heuristic, and a genuine EWMA next-touch predictor, \texttt{predictive}), the block granularity, and the prefetch lookahead. Principal settings are in Table~\ref{tab:hparams}.

One protocol detail needs stating up front. The policy grid runs chat as six concurrent sessions of 4{,}000 accesses rather than the canonical single 24{,}000-access session (the same total volume). The reason is measurability, not tuning: a single dense-attention session arrives as \emph{one} request, so there are no cross-request re-reads for a placement policy to act on, and every eviction policy produces byte-identical results (that run is preserved in \texttt{results/metrics\_all\_single\_session.csv}). The collapse is itself a finding (for one isolated long chat, eviction order does not matter), and the six-session setting is the smallest concurrency at which the policies separate.

\begin{table}[t]
\caption{Principal settings.}
\label{tab:hparams}
\centering
\resizebox{\columnwidth}{!}{%
\begin{tabular}{ll}
\toprule
Setting & Value \\
\midrule
Tiers & GPU HBM, CPU DRAM, SSD \\
Placement policy & recency / reuse-frequency / predicted-reuse / predictive (EWMA) \\
Block granularity & swept \\
Prefetch lookahead & swept \\
Baselines & GPU-only, full-CPU-offload, prefix-reuse \\
Workloads & chat, agent, document QA \\
\bottomrule
\end{tabular}%
}
\end{table}

\section{Results}
\label{sec:results}

All numbers in this section are measured by the simulator and read into the text through generated macros; the figures are drawn from the same result CSVs by \texttt{paper/make\_figures.py}. Sessions-per-GPU and cost are reported at the repository's default tier shape $1{:}8{:}64$ (working-set-to-HBM ratio $R\!\approx\!73$); the policy comparison uses the synthetic generators at their default configuration.

Table~\ref{tab:main} is the headline: tiering an HBM-only cache into HBM$+$DRAM$+$SSD multiplies the concurrent long-lived sessions a GPU sustains and divides cost per session by the same factor. That factor tracks the tier shape's capacity multiple ($1{+}8{+}64\!=\!73$), nearly independent of the workload's reuse structure, so it is a capacity-accounting result, not a placement-policy result. The placement policy shows up elsewhere, in migration traffic and TTFT (Sec.~\ref{sec:policy}).

\begin{table}[t]
\caption{Sessions per GPU and cost per session, HBM-only versus tiered (HBM$+$DRAM$+$SSD), at tier shape $1{:}8{:}64$. The gain equals the tier's capacity multiple, not a policy effect. Source: \texttt{results/ratio\_sweep.csv}.}
\label{tab:main}
\centering
\begin{tabular}{lccc}
\toprule
& Chat & Agent & Doc.\ QA \\
\midrule
Sessions/GPU, HBM-only & \mChatSessionsHBM & \mAgentSessionsHBM & \mDocqaSessionsHBM \\
Sessions/GPU, tiered & \mChatSessionsTiered & \mAgentSessionsTiered & \mDocqaSessionsTiered \\
Gain ($\times$) & \mChatSessionsGain & \mAgentSessionsGain & \mDocqaSessionsGain \\
\midrule
Cost/session, HBM-only (\$) & \mChatCostHBM & \mAgentCostHBM & \mDocqaCostHBM \\
Cost/session, tiered (\$) & \mChatCostTiered & \mAgentCostTiered & \mDocqaCostTiered \\
Reduction ($\times$) & \mChatCostSavingsFactor & \mAgentCostSavingsFactor & \mDocqaCostSavingsFactor \\
\midrule
P99 TTFT, tiered (ms) & \mChatPNinetyNineAtRefRatio & \mAgentPNinetyNineAtRefRatio & \mDocqaPNinetyNineAtRefRatio \\
\bottomrule
\end{tabular}
\end{table}

\subsection{Sessions per GPU}
\begin{figure}[t]
\centering
\includegraphics[width=\columnwidth]{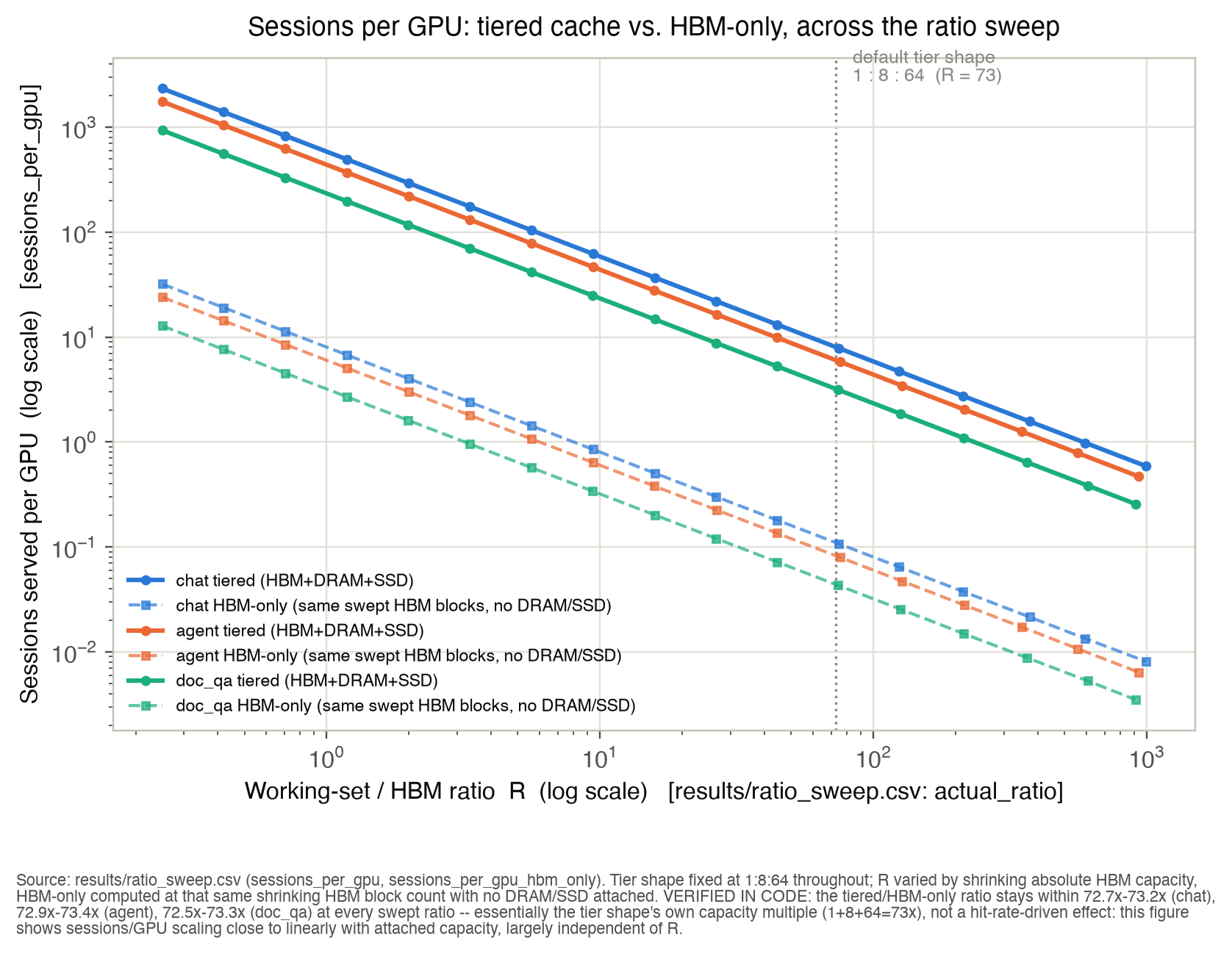}
\caption{Sessions per GPU, tiered (HBM$+$DRAM$+$SSD) versus HBM-only, across the working-set-to-HBM ratio sweep; the dotted line marks the default $1{:}8{:}64$ shape. The tiered/HBM-only ratio is flat at $\approx$\mChatSessionsGain$\times$ throughout, which is exactly the tier's capacity multiple; hit rate does not enter it. Source: \texttt{results/ratio\_sweep.csv}.}
\label{fig:savings}
\end{figure}

Figure~\ref{fig:overflow} is the motivation and Fig.~\ref{fig:savings} the payoff. Across the whole tier-ratio sweep the tiered-to-HBM-only sessions-per-GPU ratio stays between \mChatSessionsGainMin$\times$ and \mChatSessionsGainMax$\times$ for chat (and similarly for the other workloads), tracking the tier's capacity multiple rather than the workload's reuse. Adding an order of magnitude of DRAM and SSD capacity fits an order of magnitude more sessions; a good placement policy adds nothing to this count, and its job is instead to protect the latency at which those sessions are served.

\subsection{Hit rate across tiers}
\begin{figure}[t]
\centering
\includegraphics[width=\columnwidth]{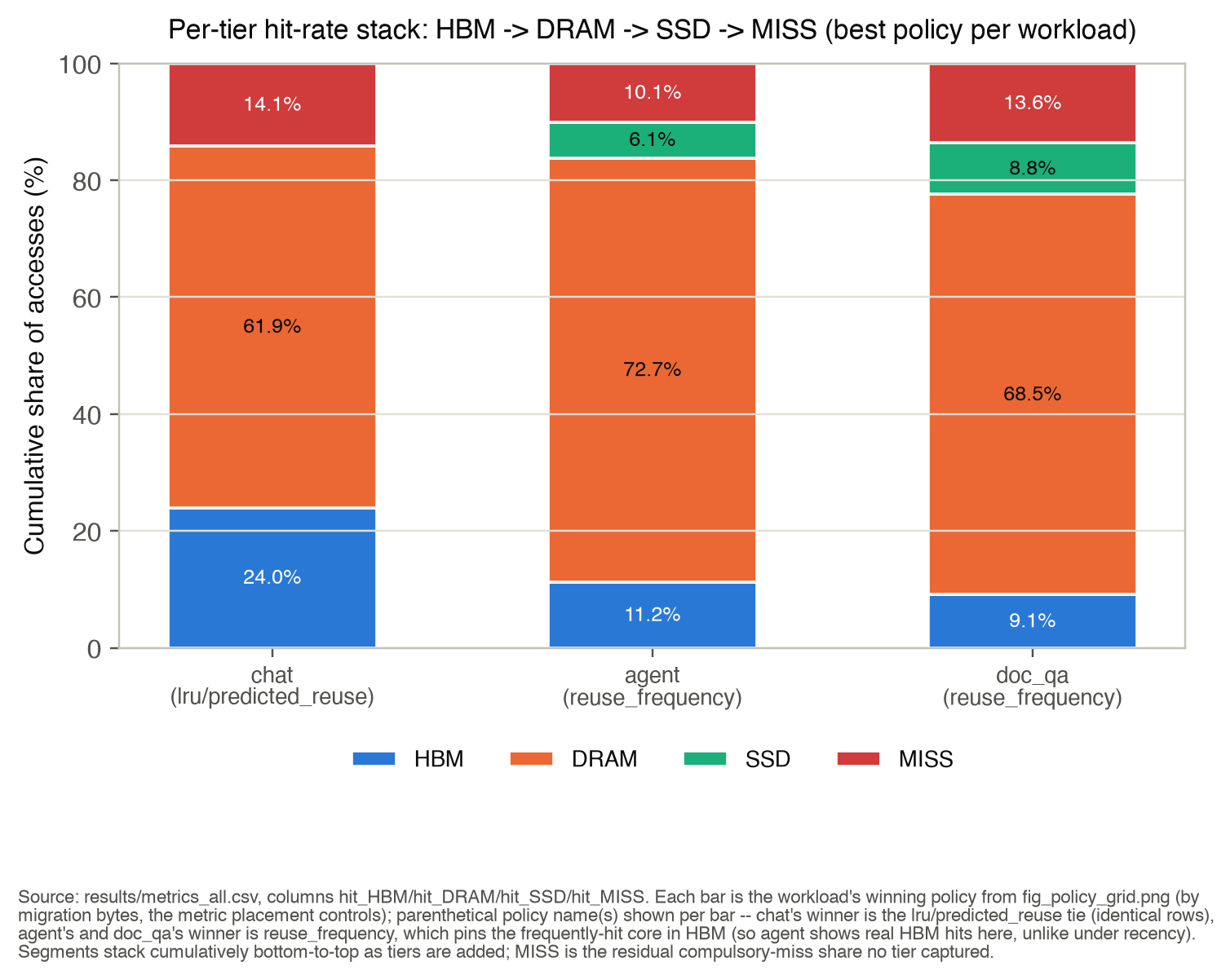}
\caption{Per-tier hit rate (HBM, DRAM, SSD, and the recomputed miss), by workload, for the winning policy. DRAM absorbs most of what overflows HBM; SSD is a shallow backstop. Source: \texttt{results/metrics\_all.csv}.}
\label{fig:hitrate}
\end{figure}

Adding DRAM and SSD changes where accesses land, and Fig.~\ref{fig:hitrate} reports the split. Under the winning policy, DRAM is the workhorse tier: chat serves \mChatHitHbm\%\ from HBM and \mChatHitDram\%\ from DRAM with \mChatHitSsd\%\ reaching SSD; agent serves \mAgentHitDram\%\ from DRAM; document QA \mDocqaHitDram\%. SSD acts as a shallow backstop (\mDocqaHitSsd\%\ on document QA, \mAgentHitSsd\%\ on agent) rather than a frequently-hit tier, and the residual miss rate that must be recomputed stays near \mChatHitMiss{} to \mDocqaHitMiss\%. So it is DRAM that carries the capacity win, at near-HBM latency; SSD earns its place on the capacity and cost it adds, not on the hits it serves.

\subsection{Which policy wins where}
\label{sec:policy}
Figure~\ref{fig:policygrid} is the core policy guide, and it carries one methodological correction: at batch one, decode is compute-bound, so effective throughput is flat across policies to within a fraction of a percent (chat is \mChatBestThroughput\ tok/s regardless of policy). Throughput is therefore the wrong lens. The placement policy controls PCIe migration traffic, and there the policies separate cleanly. For chat, recency (\texttt{lru}, and its identical twin \texttt{predicted\_reuse}) moves \mChatMigrBestGB\ GB against reuse-frequency's \mChatMigrWorstGB\ GB, a \mChatMigrWinFactor$\times$ difference, and also wins TTFT, though this is the \emph{concurrent}-chat number. For the roadmap's own scenario, one isolated long chat, the protocol disclosure in Sec.~IV applies. Every eviction policy is byte-identical there, so the roadmap's ``cheap LRU tracks it well'' is right for an unexpected reason. When eviction order cannot matter, the cheapest policy is automatically the correct one, and the binding constraint is capacity (Table~\ref{tab:main}), just as the roadmap's ``stresses: HBM capacity'' anticipated. For agent and document QA the order reverses, and reuse-frequency wins, at \mAgentMigrBestGB\ GB (\mAgentMigrWinFactor$\times$) and \mDocqaMigrBestGB\ GB (\mDocqaMigrWinFactor$\times$) respectively. So two of the roadmap's three policy recommendations hold on the bandwidth axis, chat favouring recency and document QA reuse-frequency.

The third does not hold as written: the shipped predicted-reuse policy is byte-\mPredReuseEqualsLru{} to recency (verified in code, not assumed), so ``predicted-reuse for agent loops'' is recency under another name, and on the agent workload recency is narrowly beaten by reuse-frequency. The grid therefore also includes \texttt{predictive}, a \emph{genuine} predictor built on an EWMA next-touch estimator that provably differs from recency, to test the roadmap's underlying premise that ``recent-context access is predictable'' with a policy that actually predicts. The premise survives, but the recommendation still does not: \texttt{predictive} finishes second on both of the workloads the roadmap's reasoning targets, at \mAgentPredictiveMigrGB\ GB against reuse-frequency's \mAgentMigrBestGB\ GB on agent and \mDocqaPredictiveMigrGB\ GB against \mDocqaMigrBestGB\ GB on document QA (a near-tie), and it trails badly on chat (\mChatPredictiveMigrGB\ GB), whose recency structure it has nothing to predict. The prefetch-grid size sweep (Sec.~\ref{sec:prefetch}) shows \texttt{predictive} taking first on agent at exactly one intermediate cache size, so prediction pays only in a narrow capacity window; at the default size, keeping the frequently-reused core resident beats predicting when each block returns. Table~\ref{tab:decision} states each recommendation beside its measured verdict.

\begin{table}[t]
\caption{The roadmap's practitioner decision table (left) beside the measured verdict (right). ``Holds'' is judged on PCIe migration traffic, the quantity placement controls at batch one.}
\label{tab:decision}
\centering
\footnotesize
\begin{tabular}{p{1.35cm}p{2.4cm}p{3.3cm}}
\toprule
Workload & Recommended & Measured verdict \\
\midrule
Document QA & Reuse-frequency $+$ prefetch & \textbf{Policy holds} (\mDocqaMigrWinFactor$\times$ less traffic); prefetch does not pay (Sec.~\ref{sec:prefetch}) \\
Agent loop & Predicted-reuse $+$ prefetch & \textbf{Does not hold as written}: predicted-reuse $\equiv$ recency; a genuine EWMA predictor finishes second; reuse-frequency wins here; prefetch does not pay \\
Long chat & Recency (LRU) & \textbf{Holds, scenario-dependent}: for the roadmap's own case (one isolated long chat), every eviction policy is byte-identical, so cheap LRU is correct by default and capacity is the only lever; under concurrent chats recency genuinely wins (\mChatMigrWinFactor$\times$ less traffic, best TTFT) \\
Tight HBM & Deep SSD $+$ prefetch & Sessions/GPU win holds (\mEtenSessionsGain$\times$, Sec.~\ref{sec:capacity-slo}); deep SSD matches the recompute SLO on document QA but does not beat it on latency, and never meets it on chat/agent; prefetch does not pay \\
Capacity & Cost-latency frontier & A genuine frontier exists (Sec.~\ref{sec:frontier}): it leans SSD-heavy, spanning \mFrontierCostSpan$\times$ in cost for at most \mFrontierMaxLatSpanPct\% latency \\
\bottomrule
\end{tabular}
\end{table}

\subsection{Where the latency goes}
\begin{figure}[t]
\centering
\includegraphics[width=\columnwidth]{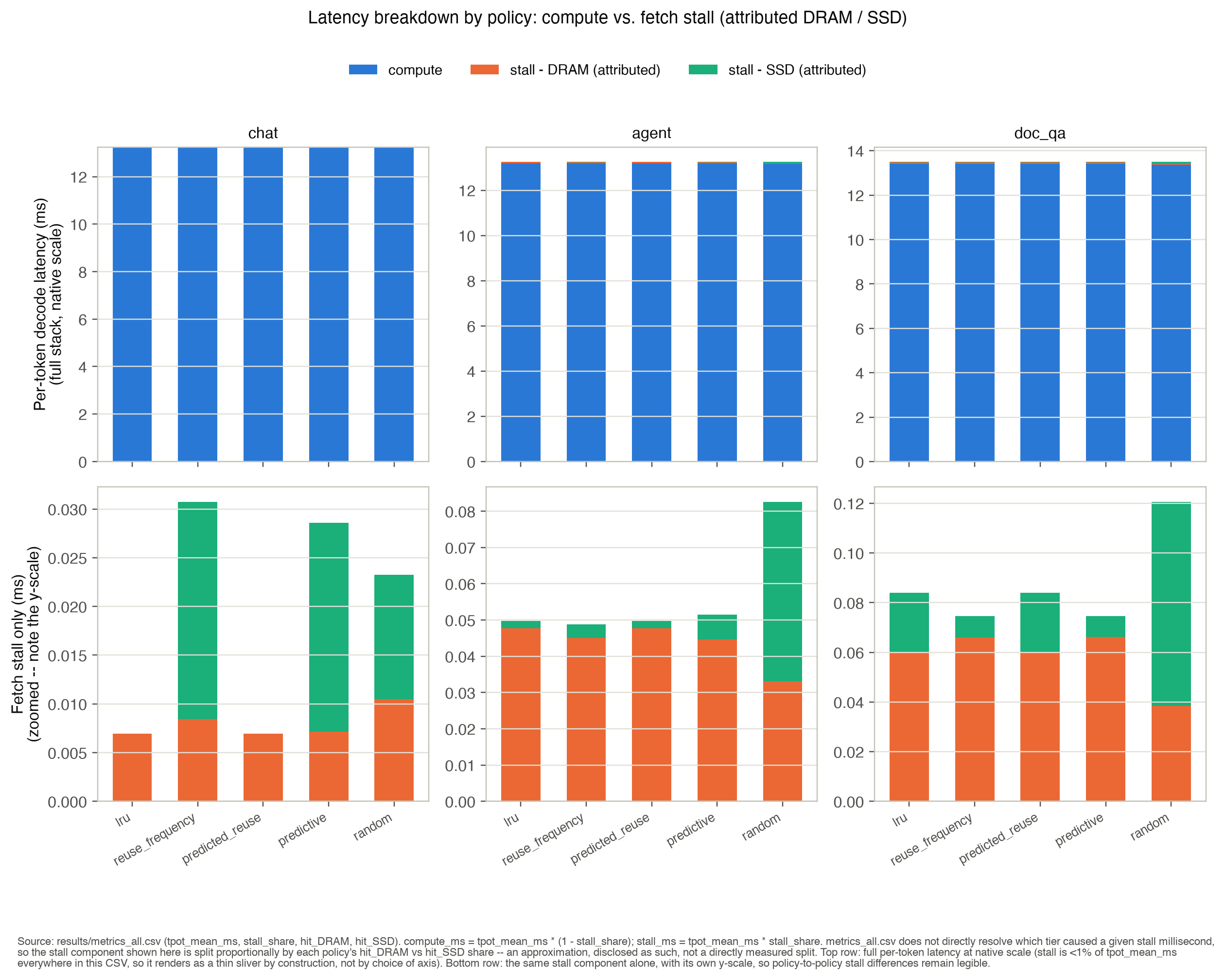}
\caption{Per-token decode latency split into compute and fetch stall, by policy; the DRAM/SSD share of the stall is attributed by hit rate (approximate). Compute dominates and stall is $<$1\% of TPOT, so tiered fetches are hidden behind compute. Source: \texttt{results/metrics\_all.csv}.}
\label{fig:latency}
\end{figure}

The latency breakdown in Fig.~\ref{fig:latency} splits per-token decode into compute and fetch stall. Compute dominates, at roughly \mChatLruComputeMs\ ms/token across policies, with fetch stall under one percent of TPOT on every workload. Two things follow. First, the tiered fetches are effectively hidden behind compute at this scale, which is the good news for the capacity win: sessions resume without a visible per-token penalty. Second, because stall is so small, TPOT is a poor discriminator between policies; the policy differences that matter live in TTFT (prefill on resume) and in the migration bandwidth of Fig.~\ref{fig:policygrid}, not here. The DRAM-versus-SSD split of the stall shown in the figure is attributed by hit share and is approximate: the simulator's per-token stall total does not record which tier caused a given stalled millisecond.

\subsection{Prefetch and the reuse structure}
\label{sec:prefetch}
The roadmap pairs each policy with ``$+$ prefetch,'' so we measured prefetch as written: for each workload we ran the policy grid a third time over a range of cache sizes with three prefetch arms (none, a causal predictor that uses only past accesses, and a future-reading oracle that is an upper bound no real system can build) on a bandwidth-charged model where a speculative fetch competes with demand traffic on the same link. The result is uniformly negative. Across the whole grid the causal predictor captures a positive share of the oracle's achievable gain in \emph{none} of the cells, and the oracle itself, despite seeing the future, is slower than not prefetching in over half of them: on a contended link a speculative fetch delays the demand fetch a request is waiting on, and that cost exceeds the stall it removes whenever the workload's reuse is not tightly predictable. Prefetch is thus the one part of the recommendation the data most clearly refutes at this scale; a moderate lookahead adds stalls here instead of removing them. This does not contradict systems where prefetch helps, since those overlap transfer with abundant spare bandwidth, but under a link the demand path is already using, ``$+$ prefetch'' is not free, and here it is not worth its bandwidth.

\subsection{The cost-latency frontier}
\label{sec:frontier}
\begin{figure}[t]
\centering
\includegraphics[width=\columnwidth]{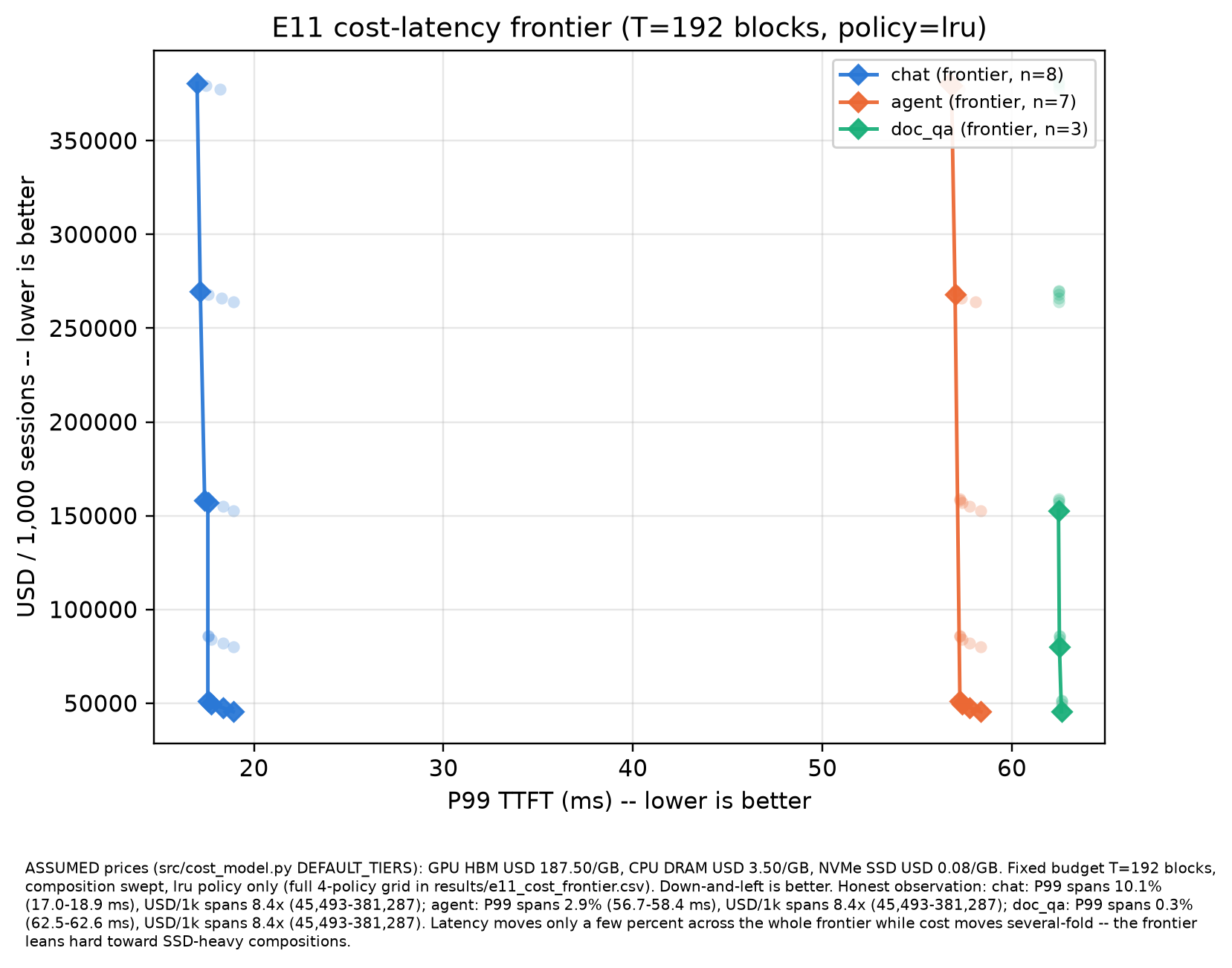}
\caption{The real cost-latency frontier (\texttt{lru} policy, one colour per workload), sweeping every (HBM, DRAM, SSD) composition at a fixed budget $T=\mFrontierBudgetT$ blocks; frontier points connected, dominated points faded. Cost per 1k sessions is a pure function of composition (verified in code, independent of workload/policy) and spans \mFrontierCostSpan$\times$; P99 TTFT moves at most \mFrontierMaxLatSpanPct\% of its ceiling across that same span, so the frontier is nearly vertical and leans toward the cheap, SSD-heavy end. Source: \texttt{results/e11\_cost\_frontier.csv}.}
\label{fig:pareto}
\end{figure}

The ratio sweep above holds tier \emph{shape} fixed at $1{:}8{:}64$ and varies only the working set; it cannot show what happens when the same fixed budget is \emph{re-shaped} across HBM, DRAM, and SSD. E11 asks that question directly: at a fixed budget $T=\mFrontierBudgetT$ blocks, we sweep every (HBM, DRAM, SSD) composition that sums to $T$, for all 3 workloads $\times$ 4 policies, and record cost and P99 TTFT at each of the 24 compositions per group. Figure~\ref{fig:pareto} plots the resulting frontier for the \texttt{lru} policy.

There is a real trade-off here, and it is a lopsided one. Cost per 1k sessions (\texttt{usd\_per\_1k\_sessions}) is, we verify in code, a pure function of the composition alone (it does not depend on workload or policy), so every one of the 12 (workload, policy) groups sweeps the identical \mFrontierCostSpan$\times$ cost range (from \$\mFrontierCostLow\ to \$\mFrontierCostHigh\ per 1k sessions). Against that \mFrontierCostSpan$\times$ cost span, P99 TTFT moves by at most \mFrontierMaxLatSpanPct\% of its own ceiling across the whole sweep (chat's \texttt{lru} and \texttt{predicted\_reuse} groups see the largest movement, \mFrontierChatLatSpanPct\%; several groups move under one percent). The frontier is therefore nearly vertical: dollars buy almost nothing in latency here, so the frontier leans hard toward the SSD-heavy end. The cheapest composition on the frontier has HBM/DRAM/SSD block counts \mFrontierCheapComposition, and it is the cheapest point in every one of the 12 groups, again checked in code; on latency the choice is nearly free. For deployment, pick the cheap, SSD-heavy composition; the HBM-heavy end of the frontier buys almost no latency for a large multiple of the price. As with the rest of this paper, the \$/GB figures behind this trade (\$187.50 / \$3.50 / \$0.08 for HBM/DRAM/SSD) are assumptions carried from \texttt{cost\_model.py}'s default tiers, not a market quote; the shape of the trade depends only on their ratio.

\subsection{Capacity under a latency SLO: how deep can SSD go?}
\label{sec:capacity-slo}
E11's frontier says cost is nearly free to minimise on latency; E10 asks the complementary question at the deployment end of that frontier: fixing a tight HBM budget (\mEtenTightHbm\ blocks) and DRAM (\mEtenDram\ blocks), how deep can SSD go before it stops being able to serve at an acceptable tail latency? E10 is a 1-D slice of E11's frontier, walking only the SSD axis at one fixed (HBM, DRAM) point and past E11's own total-budget cap, so its operating points land directly on E11's grid.

The SLO we hold it to is each workload's own \texttt{gpu\_only} (HBM-only, recompute-on-miss) baseline P99 TTFT, which avoids picking an arbitrary millisecond figure: \mSloChat\ ms for chat, \mSloAgent\ ms for agent, \mSloDocqa\ ms for document QA. Deepening SSD from its shallowest to its deepest swept point does maximise sessions per GPU, by \mEtenSessionsGain$\times$ at every workload; the gain is capacity-driven and identical across workloads, as expected from a fixed HBM/DRAM point. But that gain is not free on latency, and it is not uniformly enough: \mEtenNeverMeetSlo. Document QA is the one workload where SSD fetching is fast enough to match, though not beat, the recompute baseline: it first meets the SLO at SSD depth \mEtenDocqaCrossDepth\ blocks, and even at the deepest swept point the headroom against the SLO is \mEtenDocqaHeadroom\ ms, exactly on the line, with a mean stall of \mEtenDocqaStall\ ms still being paid at that point. So row 4's ``known stall cost'' is real and it is small on document QA, where deep SSD is roughly latency-neutral against recompute; on chat and agent, deep SSD never overtakes recompute at all in the swept range, so the sessions-per-GPU win there is bought at a latency premium the SLO does not accept.

\begin{figure}[t]
\centering
\includegraphics[width=\columnwidth]{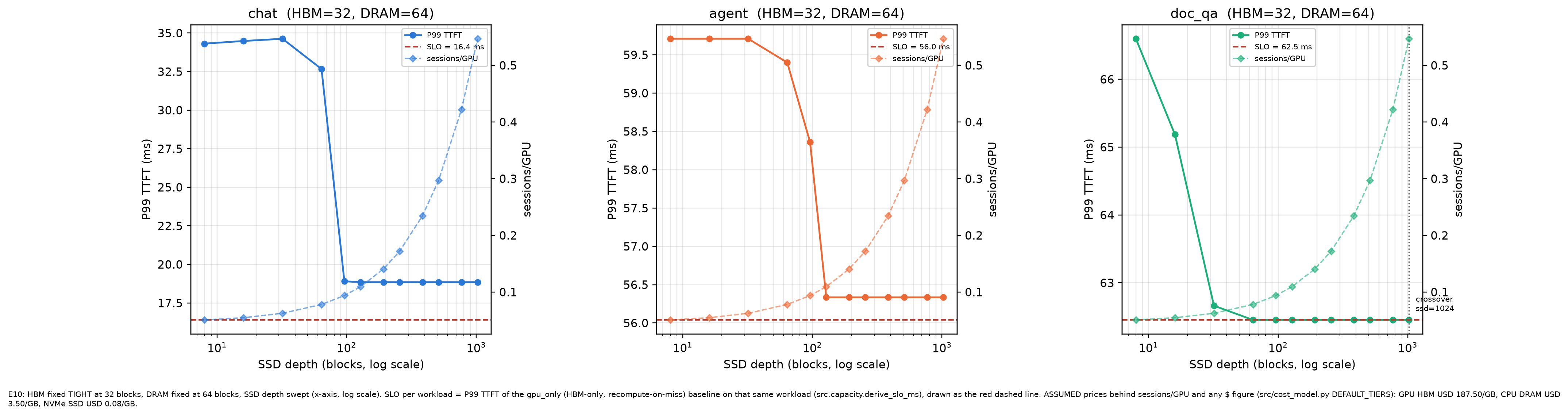}
\caption{Tight HBM (\mEtenTightHbm\ blocks) and fixed DRAM (\mEtenDram\ blocks), SSD depth swept 8 to 1024 blocks: P99 TTFT and sessions/GPU vs.\ SSD depth (log scale), one panel per workload. The SLO (dashed) is each workload's own \texttt{gpu\_only} baseline P99 TTFT; the crossover (dotted), where present, marks the shallowest SSD depth that meets it. Chat and agent never cross; document QA crosses at depth \mEtenDocqaCrossDepth\ and matches the SLO with \mEtenDocqaHeadroom\ ms headroom at the deepest point. Source: \texttt{results/e10\_capacity.csv}.}
\label{fig:capacity-slo}
\end{figure}

\section{Discussion and Limitations}
The clearest finding is a separation of concerns. The capacity and cost wins are large (\mChatSessionsGain$\times$ more sessions per GPU, \mChatCostSavingsFactor$\times$ lower cost) but they are \emph{mechanism} wins: they follow from attaching cheap DRAM and SSD, and any correct placement realises them. The placement \emph{policy} is a second-order, bandwidth-and-latency decision, and there the guide is workload-specific in the direction the roadmap anticipated for two of three cases: recency for chat, reuse-frequency for document QA. The two parts that fail are instructive. ``Predicted-reuse'' is, as shipped, byte-identical to recency, so it cannot be a distinct winner. We therefore added a policy built on an EWMA next-touch predictor. It does differ from recency, and it still finishes second to reuse-frequency on the agent loop at the default cache size; prediction's value is confined to a narrow capacity window, and a predictor that wins broadly remains open. ``$+$ prefetch'' fails because our model charges speculative fetches on the same contended link as demand traffic; on that model prefetch removes stalls only when spare bandwidth exists, which at batch one it largely does not.

Several limitations bound these claims. The policy comparison runs on synthetic generators whose reuse structure is, by construction, favourable to the policy that matches it; on real conversation traces reuse is lower and the margins compress, so the synthetic separations are upper bounds on how cleanly the policies divide. The simulator runs at batch one on a single GPU, where decode is compute-bound and fetch stall is under one percent of TPOT; at larger batch or with faster decode kernels the stall share, and hence the value of good placement and of prefetch, would rise. Sessions-per-GPU and cost are capacity-model outputs from fixed tier prices, exact given those inputs but only as representative as the prices. The DRAM/SSD split of decode stall is attributed by hit share, not directly measured. And placement composes with KV quantization and eviction, which shrink blocks before they are placed; we hold those fixed to isolate placement.

\section{Conclusion}
We argued that the mechanism for extending the KV cache beyond GPU memory is the easy half and the placement policy the hard one, and built a calibrated tiered simulator over GPU HBM, CPU DRAM, and SSD to test that claim. The measured answer refines it. The mechanism delivers the headline: tiering multiplies sessions per GPU by \mChatSessionsGain$\times$ and lowers cost per session by \mChatCostSavingsFactor$\times$, a capacity result independent of policy. The policy is a bandwidth decision, and the recipe that holds is compact: recency for concurrent chat serving (for one isolated long chat, every policy is identical and cheap LRU is right by default), reuse-frequency for repeated document QA, both judged on the PCIe traffic placement actually controls. The two elements of the original recommendation that do not survive (predicted-reuse, which as implemented is recency, and $+$ prefetch, which on a contended link costs more than it saves) are as useful to report as the parts that hold. The EWMA predictor we added to close the first gap differs from recency yet still loses to reuse-frequency at the default cache size, so the next work is sharper than ``build a predictor'': a predictor that wins outside a narrow capacity window, and a prefetch model with spare bandwidth to hide behind.

\section*{Acknowledgments}
The authors thank Vizuara AI Labs for mentorship and computational resources.

\end{document}